\documentclass[11pt]{article}
\usepackage{textgreek}

\usepackage[preprint]{acl}

\usepackage{times}
\usepackage{latexsym}

\usepackage{float}

\usepackage[T1]{fontenc}

\usepackage[utf8]{inputenc}

\usepackage{microtype}

\usepackage{inconsolata}

\usepackage{graphicx}

\usepackage{booktabs}       
\usepackage{multirow}       
\usepackage{amsmath}        
\usepackage{amssymb}        
\usepackage{enumitem}       
\usepackage{xcolor}         
\usepackage{subcaption}     

\usepackage{titlesec}
\titlespacing*{\section}{0pt}{2.0ex plus 1ex minus .2ex}{1.0ex plus .2ex}
\titlespacing*{\subsection}{0pt}{1.5ex plus 1ex minus .2ex}{0.8ex plus .2ex}
\titlespacing*{\subsubsection}{0pt}{1.0ex plus 0.5ex minus .2ex}{0.5ex plus .2ex}

\setlist{noitemsep, topsep=2pt, parsep=0pt, partopsep=0pt}
\setlist[enumerate]{leftmargin=1.5em}
\setlist[itemize]{leftmargin=1.5em}

\newcommand{\squishenum}{
 \begin{enumerate}
  \setlength{\itemsep}{0pt}
  \setlength{\parsep}{3pt}
  \setlength{\topsep}{3pt}
  \setlength{\partopsep}{0pt}
  \setlength{\leftmargin}{1.8em}
 }
\newcommand{\squisheend}{\end{enumerate}}

\newcommand{\squishlist}{
 \begin{list}{$\bullet$}
  { \setlength{\itemsep}{0pt}
     \setlength{\parsep}{3pt}
     \setlength{\topsep}{3pt}
     \setlength{\partopsep}{0pt}
     \setlength{\leftmargin}{1.5em}
     \setlength{\labelwidth}{1em}
     \setlength{\labelsep}{0.5em} } }

\newcommand{\squishlisttwo}{
 \begin{list}{$\bullet$}
  { \setlength{\itemsep}{0pt}
    \setlength{\parsep}{0pt}
    \setlength{\topsep}{0pt}
    \setlength{\partopsep}{0pt}
    \setlength{\leftmargin}{2em}
    \setlength{\labelwidth}{1.5em}
    \setlength{\labelsep}{0.5em} } }

\newcommand{\squishend}{
  \end{list}  }

\title{Can Embedding Similarity Predict Cross-Lingual Transfer? \\ A Systematic Study on African Languages}

\author{Tewodros Kederalah Idris\textsuperscript{1} \quad Prasenjit Mitra\textsuperscript{1} \quad 
Roald Eiselen\textsuperscript{2} \\
  \textsuperscript{1}Carnegie Mellon University Africa \\
  \textsuperscript{2}North-West University \\
  \texttt{tidris@andrew.cmu.edu, prasenjm@andrew.cmu.edu, roald.eiselen@nwu.ac.za}}

\begin{document}
\maketitle

\begin{abstract}

Cross-lingual transfer is essential for building NLP systems for low-resource African languages, but practitioners lack reliable methods for selecting source languages. We systematically evaluate five embedding similarity metrics across 816 transfer experiments spanning three NLP tasks, three African-centric multilingual models, and 12 languages from four language families. We find that cosine gap and retrieval-based metrics (P@1, CSLS) reliably predict transfer success ($\rho = 0.4$--$0.6$), while CKA shows negligible predictive power ($\rho \approx 0.1$). Critically, correlation signs reverse when pooling across models (Simpson's Paradox), so practitioners must validate per-model. Embedding metrics achieve comparable predictive power to URIEL linguistic typology. Our results provide concrete guidance for source language selection and highlight the importance of model-specific analysis.

\end{abstract}

\section{Introduction}

Cross-lingual transfer enables NLP systems to process languages that lack labeled training data. The approach is simple: train a model on a high-resource source language, then apply it to a low-resource target language. For many African languages, where labeled data is scarce or nonexistent, this is often the only viable path to building working NLP systems \citep{adelani-etal-2021-masakhaner, adelani-etal-2022-masakhaner, dione-etal-2023-masakhapos}.

The choice of source language matters. \citet{adelani-etal-2022-masakhaner} 
showed that selecting the right source language can improve
performance by up to 14 F1 points compared to using English for African language NER. However, we lack reliable methods for making this selection. 
We choose languages from the same family, consult typological databases such as URIEL \citep{littell-etal-2017-uriel}, or simply use the largest available dataset. Each of these heuristics
is useful, but none consistently identifies the best source language for different tasks and models \citep{rice-etal-2025-typology, litschko-etal-2020-evaluating}.

A more direct approach would be to ask the model itself. Multilingual encoders represent sentences from different languages in a shared embedding space.
\citet{pires-etal-2019-multilingual} have shown that the degree of cross-lingual alignment in these representations correlates with transfer performance. If two languages have aligned representations, a classifier trained on one should generalize to the other. This suggests a practical strategy: compute the similarity between source and target language embeddings, and use this similarity to predict transfer performance.

However, the relationship between embedding similarity and transfer success is not straightforward. The structural properties of multilingual embeddings, such as anisotropy, where all vectors cluster in a narrow cone \citep{ethayarajh-2019-contextual, rajaee-pilehvar-2021-cluster} can distort standard similarity measures. Different metrics may capture different aspects of the cross-lingual structure and their effectiveness may vary between models and tasks. Although embedding similarity metrics are commonly used in cross-lingual retrieval \citep{conneau-etal-2018-xnli, artetxe-schwenk-2019-massively}, it remains unclear whether they reliably predict downstream transfer performance --especially for African languages.

This question is particularly pressing for African languages, where typological databases have limited coverage \citep{adelani-etal-2022-thousand, adebara-abdul-mageed-2022-towards} and domain mismatch may undermine metric reliability \citep{hedderich-etal-2020-transfer}.

We conduct a systematic empirical study to investigate these questions. We evaluate five embedding similarity metrics across three multilingual models trained on African languages: AfriBERTa \citep{ogueji-etal-2021-small}, AfroXLM-R \citep{alabi-etal-2022-adapting}, and Serengeti \citep{adebara-etal-2023-serengeti}. The metrics span three categories: magnitude-based (cosine, cosine gap), retrieval-based (P@1, CSLS), and representation-based (CKA). We compute these metrics for all pairs of 12 African languages spanning four language families, then measure actual transfer performance on three downstream tasks: named entity recognition, part-of-speech tagging, and sentiment analysis for a total of 816 transfer experiments.

Our analysis addresses three questions: (1) which metrics predict transfer, (2) whether results generalize across models, and (3) how embedding metrics compare to linguistic typology. We find that cosine gap and retrieval-based metrics (P@1, CSLS) reliably predict transfer ($\rho = 0.4$--$0.6$), while CKA shows negligible predictive power ($\rho \approx 0.1$). Critically, correlation signs reverse when pooling across models (Simpson's Paradox)~\cite{simpson1951interpretation}, so practitioners must validate per-model. Embedding metrics achieve comparable predictive power to URIEL genetic distance (mean $|\rho|$ = 0.50 for cosine\_gap vs. 0.44 for URIEL across NER and POS; see Table 7), and metrics predict transfer well for NER and POS but fail for sentiment analysis due to domain mismatch. These results provide concrete guidance for source language selection while highlighting that model-specific validation remains essential.

\section{Related Work}
\label{sec:related-work}

\subsection{Cross-Lingual Transfer and Source Language Selection}



Cross-lingual transfer leverages labeled data from high-resource languages to build NLP systems for low-resource languages \citep{pires-etal-2019-multilingual, conneau-etal-2020-unsupervised}. Transfer works best between typologically similar languages \citep{pires-etal-2019-multilingual}, and the choice of source language significantly impacts performance: \citet{adelani-etal-2022-masakhaner} showed up to 14 F1 point differences across source languages for African NER, while \citet{turc-etal-2021-revisiting} found that German and Russian often transfer more effectively than English for classification tasks.

Several studies have attempted to predict transfer performance. \citet{lauscher-etal-2020-zero} found that linguistic proximity predicts transfer for low-level tasks (POS tagging, dependency parsing, NER), with pretraining corpus size also playing a significant role. \citet{lin-etal-2019-choosing} proposed LANGRANK, combining URIEL typological features \citep{littell-etal-2017-uriel} with dataset statistics, finding that no single feature reliably identifies the best source across tasks. However, these approaches rely on static linguistic features that do not capture model-specific cross-lingual structure learned during pretraining.


\subsection{Embedding Similarity Metrics}

The cosine similarity between sentence embeddings is widely used but suffers from anisotropy \citep{ethayarajh-2019-contextual, gao-etal-2021-simcse}. Retrieval-based metrics such as P@1 and CSLS \citep{conneau-etal-2017-word} address this through nearest-neighbor retrieval. CKA \citep{kornblith2019similarity} measures structural similarity, although \citet{del-fishel-2022-cross} argue that its assumptions align poorly with cross-lingual transfer.

The key question is whether these metrics predict transfer performance. Recent work suggests that they can: \citet{kargaran-etal-2025-mexa} report Pearson's correlation $r = 0.90$ with downstream performance, \citet{wang-etal-2024-probing-emergence} found neuron overlap correlates with zero-shot transfer, and \citet{zhao-etal-2023-joint} showed matrix factorization outputs associated with cross-lingual task performance. However, \citet{ravisankar-etal-2025-map} caution that alignment is ``a necessary but insufficient condition,'' often failing at sample-level prediction.

\subsection{Linguistic Typology for Transfer Prediction}




Beyond learned ranking approaches, typological features can directly predict transfer. \citet{muller-etal-2023-languages} conducted a systematic analysis of mT5 across 90 language pairs, finding that syntactic, morphological, and phonological similarity are better predictors than lexical similarity. However, typological databases have incomplete coverage for many African languages \citep{adelani-etal-2022-thousand, adebara-abdul-mageed-2022-towards}, motivating investigation of embedding-based metrics that can be computed for any language pair covered by a multilingual model.

\subsection{African Language NLP}

African languages present unique challenges for NLP: over 2,000 languages with limited digital resources, morphological complexity (particularly in Bantu languages), and significant typological variation across language families ~\cite{creissels-2019-morphology-niger-congo, hyman-etal-2019-niger-congo}


\citet{kargaran-etal-2025-mexa} reported that their cross-lingual alignment metric (MEXA) achieves an average Pearson correlation of $r = 0.90$ with downstream task performance across languages, while \citet{muller-etal-2021-first} found that CKA-based cross-lingual similarity correlates with transfer gap in mBERT. Whether such strong correlations hold for African languages, which exhibit distinct typological properties and are underrepresented in model pretraining, remains untested.

\section{Methods}
\label{sec:methods}

We evaluate five embedding similarity metrics from three families: magnitude-based, retrieval-based, and structural. Our goal is to determine which metrics best predict zero-shot cross-lingual transfer performance, where a model is fine-tuned on source language data and evaluated directly on a target language without any target language training. All metrics are computed on parallel sentences from FLORES-200 \citep{nllb2024}.

\subsection{Embedding Similarity Metrics}
{\bf Sentence Encoding.}
We compute sentence embeddings via mean pooling over final-layer token representations, following standard practice \citep{conneau-etal-2020-emerging, muller-etal-2021-first}, and L2-normalize the results. We use the FLORES-200 devtest split (1,012 parallel sentences per language) consistently across all three models.

{\bf Notation.}
Let $S = \{s_1, \ldots, s_N\}$ and $T = \{t_1, \ldots, t_N\}$ denote L2-normalized sentence embeddings for source and target languages respectively, where $(s_i, t_i)$ are translations of the same sentence. Let $\mathbf{M} \in \mathbb{R}^{N \times N}$ denote the similarity matrix where $M_{ij} = s_i^\top t_j$.

\subsubsection{Magnitude-Based Metrics}

{\bf Cosine Similarity.}
We compute the mean cosine similarity between aligned sentence pairs.
Since embeddings are L2-normalized, this equals the mean of diagonal entries of $\mathbf{M}$.

{\bf Cosine Gap.}
Raw cosine similarity can be misleading due to anisotropy, where embeddings concentrate in a narrow cone of the vector space \citep{ethayarajh-2019-contextual, gao-etal-2021-simcse}. 
\citet{kargaran-etal-2025-mexa} show that comparing parallel and non-parallel sentence similarities better captures cross-lingual alignment. We compute the gap between aligned and baseline similarity:
\begin{equation}
\text{cosine\_gap} = \text{cosine\_mean} - \frac{1}{N^2} \sum_{i,j} M_{ij}
\end{equation}
Here, \emph{aligned} pairs $(s_i, t_i)$ are parallel sentences (translations of each other), while the baseline averages over all $N^2$ pairs including \emph{misaligned} pairs $(s_i, t_j)$ where $i \neq j$. A higher gap indicates that true translation pairs are more distinguishable from arbitrary sentence pairings.

\subsubsection{Retrieval-Based Metrics}

{\bf Precision at 1 (P@1):} is the fraction of source sentences whose nearest neighbor in the target embedding space is the correct translation.
This metric is asymmetric; we compute both $\text{P@1}(S \rightarrow T)$ and $\text{P@1}(T \rightarrow S)$, reporting the source-to-target direction to match the transfer learning setting.

{\bf Cross-domain Similarity Local Scaling (CSLS).}
CSLS addresses the hubness problem, where some embeddings become nearest neighbors to many others regardless of true semantic similarity \citep{conneau-etal-2017-word}. For each embedding, we compute the mean similarity to its $k$ nearest neighbors as a measure of its ``hubness'':
\begin{equation}
r_T(s_i) = \frac{1}{k} \sum_{t_j \in \mathcal{N}_k(s_i)} s_i^\top t_j
\end{equation}
where $\mathcal{N}_k(s_i)$ denotes the $k$ nearest neighbors of $s_i$ in $T$. The CSLS-corrected similarity is:
\begin{equation}
\text{CSLS}(s_i, t_j) = 2 \cdot s_i^\top t_j - r_T(s_i) - r_S(t_j)
\end{equation}
We use $k=10$ following standard practice and report the mean CSLS score across aligned pairs.

\subsubsection{Structural Metrics}

{\bf Centered Kernel Alignment (CKA).}
CKA measures whether two representation spaces preserve the same similarity structure among examples \citep{kornblith2019similarity}. Let $\mathbf{K} = S S^\top$ and $\mathbf{L} = T T^\top$ be the Gram matrices capturing pairwise similarities within each language. CKA is defined as:
\begin{equation}
\text{CKA}(S, T) = \frac{\text{HSIC}(\mathbf{K}, \mathbf{L})}{\sqrt{\text{HSIC}(\mathbf{K}, \mathbf{K}) \cdot \text{HSIC}(\mathbf{L}, \mathbf{L})}}
\end{equation}
where HSIC is the Hilbert-Schmidt Independence Criterion. We use the unbiased HSIC estimator \citep{song2012feature} with a linear kernel. CKA is symmetric and bounded in $[0, 1]$.

Recent work has questioned whether CKA assumptions hold for cross-lingual comparison \citep{del-fishel-2022-cross}. We include it to empirically test its predictive power for African language transfer.

{\bf Summary.}
Table~\ref{tab:metrics-summary} summarizes the five metrics. Magnitude-based metrics measure direct similarity; retrieval-based metrics measure alignment quality through nearest-neighbor retrieval; CKA measures global structural similarity.

\begin{table}[htb]
\centering
\small
\begin{tabular}{llcc}
\toprule
\textbf{Metric} & \textbf{Family} & \textbf{Addresses} & \textbf{Range} \\
\midrule
cosine\_mean & Magnitude & -- & $[-1, 1]$ \\
cosine\_gap & Magnitude & Anisotropy & $[-2, 2]$ \\
P@1 & Retrieval & -- & $[0, 1]$ \\
CSLS & Retrieval & Hubness & $(-\infty, \infty)$ \\
CKA & Structural & -- & $[0, 1]$ \\
\bottomrule
\end{tabular}
\caption{Summary of embedding similarity metrics. Higher values indicate greater similarity or better alignment for all metrics.}
\label{tab:metrics-summary}
\end{table}

\subsection{Correlation Analysis Framework}

To evaluate which metrics predict transfer performance, we compute Spearman rank correlation ($\rho$) between metric values and downstream task performance across language pairs.

{\bf Procedure.}
For each combination of model and task:
\squishlist
\item Compute all five metrics for each source-target language pair using FLORES-200 embeddings.
\item Perform zero-shot cross-lingual transfer: fine-tune on source language training data, evaluate directly on target language test data.
\item Calculate Spearman's $\rho$ between metric values and transfer performance.
\squishend
{\bf Why Spearman Correlation?}
We use Spearman and not Pearson because it measures monotonic relationships without assuming linearity, is robust to outliers, and directly addresses our goal: ranking language pairs by expected transfer performance.

{\bf Stratification Strategy.}
We analyze correlations at multiple levels of granularity:
\begin{enumerate}[leftmargin=1.5em, topsep=0.3em, itemsep=0.2em]
    \item \textbf{Primary (per-model, per-task):} Correlations computed separately for each of the 9 model-task combinations (e.g., AfriBERTa-NER with $n = 132$ pairs, AfroXLM-R-POS with $n = 110$ pairs). These are our main reported results.
    \item \textbf{Within-task pooling:} Correlations computed across all models for a single task (e.g., all NER pairs, $n = 396$). We use this to demonstrate Simpson's Paradox (\S\ref{sec:simpson}).
    \item \textbf{Fully pooled:} We do not report fully pooled correlations, as they conflate model, task, and language effects.
\end{enumerate}
Different models exhibit fundamentally different cross-lingual alignment structures due to variations in pretraining data and objectives. Pooling across models introduces confounds that can reverse correlation signs, as we demonstrate in \S\ref{sec:simpson}.


\subsection{Linguistic Baseline: URIEL Distances}
We compare embedding-based metrics against linguistic typology as a baseline predictor of transfer performance.

{\bf URIEL Database.}
We extract five distance types: genetic (language family tree distance), syntactic (word order and agreement patterns), phonological (sound system properties), inventory (phoneme repertoire overlap), and geographic (physical distance between speaker populations).

{\bf Comparison Framework.}
We compute Spearman correlations between URIEL distances and transfer performance. Since distances (unlike similarities) should correlate negatively with performance, we compare $|\rho|$ values across embedding metrics and linguistic distances. This comparison addresses whether embedding metrics provide complementary information beyond static typological features, and whether model-specific metrics outperform model-agnostic linguistic distances.

\section{Experimental Setup}
\label{sec:experiments}


\subsection{Models}
\label{sec:models}


\begin{table}[htb]
\centering
\small
\begin{tabular}{llrr}
\toprule
\textbf{Model} & \textbf{Architecture} & \textbf{Params} & \textbf{Languages} \\
\midrule
AfriBERTa & RoBERTa & 126M & 11 \\
AfroXLM-R & XLM-RoBERTa & 550M & 17+ \\
Serengeti & XLM-RoBERTa & 278M & 517+ \\
\bottomrule
\end{tabular}
\caption{Multilingual African language models. All use masked language modeling.\protect\footnotemark}
\label{tab:models}
\end{table}
\footnotetext{HuggingFace IDs: \texttt{castorini/afriberta\_large}, \texttt{Davlan/afro-xlmr-large}, \texttt{UBC-NLP/serengeti}. Code released upon publication.}

{\bf AfriBERTa} \citep{ogueji-etal-2021-small} is a small but competitive model pretrained using a RoBERTa architecture. 

{\bf AfroXLM-R} \citep{alabi-etal-2022-adapting} extends XLM-RoBERTa through multilingual adaptive fine-tuning, improving cross-lingual transfer for low-resource African languages.

{\bf Serengeti} \citep{adebara-etal-2023-serengeti} is the largest-scale African language model, covering 517+ languages with a focus on low-resource African languages.

{\bf Pretraining Language Coverage.}
AfriBERTa covers 6 of our 12 languages, AfroXLM-R covers 9, and Serengeti covers all 12 (see Table~\ref{tab:lang-coverage-app} in Appendix). We analyze the effect of pretraining inclusion on transfer performance in Section~\ref{sec:pretraining-effects}.


{\bf Architectural Homogeneity.}
All three models share the RoBERTa/XLM-RoBERTa architecture family and use the masked language modeling objective. This homogeneity enables controlled comparison: observed differences in cross-lingual alignment and transfer performance can be attributed to pretraining data distribution and language coverage rather than architectural confounds. We exclude general-purpose multilingual models (e.g., mBERT, XLM-R) to keep the study focused on African-centric models and to avoid confounding effects from different pretraining regimes and data distributions.

\subsection{Languages}
\label{sec:languages}

We evaluate on twelve African languages spanning four language families and two writing systems (Table~\ref{tab:languages}).

\begin{table}[htb]
\centering
\small
\begin{tabular}{@{}llp{1.2cm}@{}}
\toprule
\textbf{Family} & \textbf{Languages} & \textbf{Script} \\
\midrule
Niger-Congo (Bantu) & kin, lug, swa, xho, zul & Latin \\
Niger-Congo (Atlantic) & wol & Latin \\
Niger-Congo (Volta-Niger) & ibo, yor & Latin \\
Afro-Asiatic & amh, hau & Ge'ez, Latin \\
Nilo-Saharan & luo & Latin \\
Mande & bam & Latin \\
\bottomrule
\end{tabular}
\caption{Languages evaluated, grouped by family. ISO 639-3 codes: amh (Amharic), bam (Bambara), hau (Hausa), ibo (Igbo), kin (Kinyarwanda), lug (Luganda), luo (Luo), swa (Swahili), wol (Wolof), xho (Xhosa), yor (Yoruba), zul (Zulu).}
\label{tab:languages}
\end{table}
{\bf Selection Rationale.}
Language selection balances typological diversity with dataset availability across four major African language families; Amharic provides script diversity as the only non-Latin script language in our sample.
\begin{table}[htb]
\centering
\small
\begin{tabular}{llcr}
\toprule
\textbf{Task} & \textbf{Dataset} & \textbf{Languages} & \textbf{Pairs} \\
\midrule
NER & MasakhaNER 2.0 & 12 & 132 \\
POS & MasakhaPOS & 11 & 110 \\
Sentiment & AfriSenti & 6 & 30 \\
\midrule
\multicolumn{3}{l}{\textbf{Total source -> target pairs (across tasks)}} & \textbf{272} \\
\bottomrule
\end{tabular}
\caption{Downstream tasks and language coverage. Language pair counts reflect ordered source-target combinations.}
\label{tab:tasks}
\end{table}

{\bf Language Pairs.}
For directed transfer analysis, we evaluate all ordered source-target pairs excluding same-language transfer. With 12 languages, this yields a maximum of $12 \times 11 = 132$ pairs per task (varies by task and dataset availability).

\subsection{Tasks and Datasets}
\label{sec:tasks}

We evaluate on two sequence labeling tasks (NER, POS) and one sentence classification task (Sentiment), spanning different domains and annotation schemes (Table~\ref{tab:tasks}).

{\bf MasakhaNER 2.0} \citep{adelani-etal-2022-masakhaner}: 
Entities are of four types: person (PER), organization (ORG), location (LOC), and date (DATE).

{\bf MasakhaPOS} \citep{dione-etal-2023-masakhapos} provides part-of-speech annotations following the Universal Dependencies tagset. 
Amharic is unavailable.

{\bf AfriSenti} \citep{muhammad-etal-2023-afrisenti} provides sentiment annotations 
for Twitter data.
We use the six languages that appear in the dataset: Amharic, Hausa, Igbo, Kinyarwanda, Swahili, and Yoruba. 

{\bf Domain Characteristics.}
NER and POS use news-domain text aligned with FLORES-200's formal style. AfriSenti uses Twitter data with informal spelling, code-mixing, testing whether formal-text metrics predict social media transfer (\S\ref{sec:domain}).

\subsection{Training Configuration}
\label{sec:training}

{\bf Fine-tuning Setup.}
We test zero-shot cross-lingual transfer by adding a task-specific classification head on top of the pretrained encoder: a linear layer for token classification (NER, POS) or sequence classification (Sentiment).

{\bf Hyperparameters.}
We adopt the fine-tuning recipe from \citet{devlin-etal-2019-bert} with modifications for training stability from \citet{mosbach-etal-2021-stability}. We use a learning rate of $2 \times 10^{-5}$ with linear warmup and decay, batch size 16, and early stopping with patience 3. Full configuration details are in Table~\ref{tab:hyperparams-app} (Appendix).



{\bf Evaluation Metrics.}
We report entity-level F1 for NER (micro-averaged using \texttt{seqeval}, following MasakhaNER evaluation protocol), macro F1 for Sentiment, and accuracy for POS.

{\bf Experimental Scale.}
We conduct 816 training runs in total: 272 unique language pairs across 3 models. Each configuration is trained once without seed averaging. Embedding similarity metrics are deterministic given fixed pretrained weights, so only transfer performance has potential variance from fine-tuning randomness. 

Our design prioritizes breadth of evaluation (272 language pairs $\times$ 3 models) over per-configuration replication, following similar large-scale transfer studies \citep{lauscher-etal-2020-zero, turc-etal-2021-revisiting}.
\section{Results}
\label{sec:results}



\subsection{Overall Metric Performance}
\label{sec:overall}


\begin{table}[htb]
\centering
\small
\begin{tabular}{@{}lccccc@{}}
\toprule
\textbf{Metric} & \textbf{Mean $\rho$} & \textbf{Std} & \textbf{Min} & \textbf{Max} & \textbf{Sig.} \\
\midrule
cosine\_mean & 0.34 & 0.16 & 0.10 & 0.53 & 6/9 \\
cosine\_gap & 0.41 & 0.16 & 0.16 & 0.60 & 6/9 \\
P@1 & 0.40 & 0.14 & 0.20 & 0.56 & 7/9 \\
CSLS & 0.40 & 0.14 & 0.23 & 0.58 & 6/9 \\
CKA & 0.10 & 0.18 & $-$0.13 & 0.38 & 2/9 \\
\bottomrule
\end{tabular}
\caption{Summary of metric-transfer correlations across 9 conditions (3 models $\times$ 3 tasks). \textbf{Sig.} indicates the number of conditions where $p < 0.05$.}
\label{tab:overall}
\end{table}
Cosine\_gap, P@1, and CSLS achieve similar moderate correlations (mean $\rho \approx 0.40$), suggesting comparable predictive signal. However, variation across conditions is substantial; no metric achieves consistently strong correlation ($\rho > 0.60$) across all settings. 
These results suggest that embedding similarity metrics provide useful but imperfect signal for predicting cross-lingual transfer.


\subsection{Simpson's Paradox: The Importance of Stratification}
\label{sec:simpson}

\begin{table}[t]
\centering
\small
\begin{tabular}{@{}lcccc@{}}
\toprule
\textbf{Analysis} & \textbf{$\rho$} & \textbf{$n$} & \textbf{Avg cos\_gap} & \textbf{Avg F1} \\
\midrule
Pooled (all) & $-$0.18 & 396 & -- & -- \\
\midrule
AfriBERTa & +0.60 & 132 & 0.035 & 0.35 \\
AfroXLM-R & +0.37 & 132 & 0.010 & 0.55 \\
Serengeti & +0.51 & 132 & 0.004 & 0.52 \\
\bottomrule
\end{tabular}
\caption{Simpson's Paradox: pooling yields negative correlation while stratified analysis reveals positive correlations (all $p < .001$). Model averages (right columns) explain the reversal.}
\label{tab:simpson-combined}
\end{table}

When we combine all 396 NER language pairs across models, the correlation between cosine\_gap and transfer F1 is \textit{negative} ($\rho = -0.18$). However, within each model, correlations are consistently \textit{positive} ($\rho = 0.37$ to $0.60$). This is Simpson's Paradox \citep{simpson1951interpretation}: AfriBERTa has the highest mean cosine\_gap (0.035) but the lowest mean F1 (0.35), while Serengeti shows the opposite pattern. This between-model confound dominates when pooled, masking within-model relationships.

\begin{figure}[t]
    \centering
    \includegraphics[width=\columnwidth]{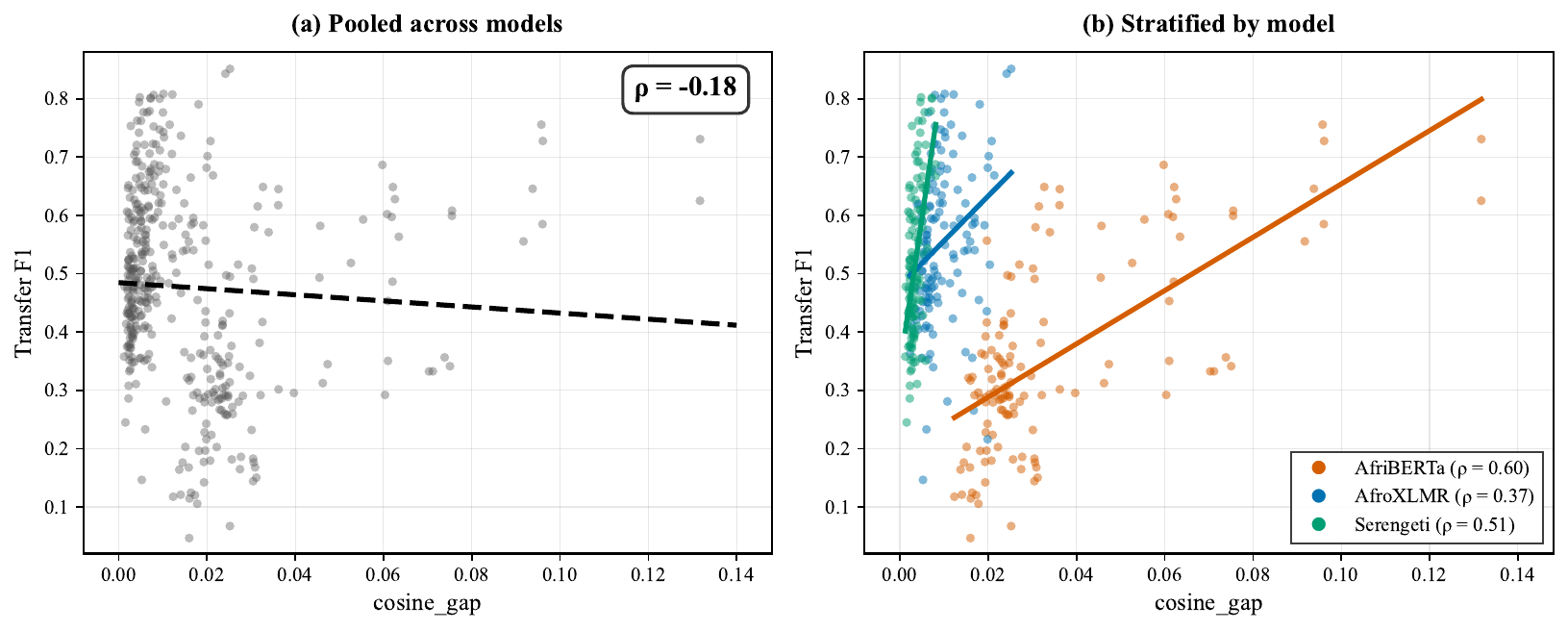}
    \caption{Simpson's Paradox in metric-transfer correlations. (a) Pooling across models suggests a -ve relationship ($\rho = -0.18$). (b) Stratifying by model reveals +ve correlations in each model. Points show all 396 NER language-pair-model combinations. Panel (a) displays points in gray with a pooled OLS regression line; Panel (b) colors points by model with separate regression lines. Legend shows per-model Spearman correlations.}
    \label{fig:simpson}
\end{figure}

Figure~\ref{fig:simpson} visualizes this effect. Panel(a) shows all language pairs with a single regression line and a weak negative relationship. Panel(b) reveals the true pattern: within each model cluster, higher cosine\_gap corresponds to higher transfer F1.

\textbf{Methodological implication:} Analyses of embedding-transfer relationships must stratify by model. Pooling across models can lead to incorrect conclusions about whether embedding metrics predict transfer performance.

\subsection{Task-Specific Results}
\label{sec:task-results}

\subsubsection{Named Entity Recognition}

NER exhibits the strongest and most consistent correlations across models (Table~\ref{tab:ner-results-app}, Appendix). For AfriBERTa, cosine\_gap achieves $\rho = 0.60$, the highest observed correlation. AfroXLM-R shows weaker correlations across all metrics, a pattern we discuss in \S\ref{sec:model-patterns}. Notably, CKA fails to achieve significant correlation for AfriBERTa and AfroXLM-R, reaching significance only for Serengeti ($\rho = 0.38$).

\subsubsection{Part-of-Speech Tagging}

POS tagging shows the most consistent pattern across models (Table~\ref{tab:pos-results-app}, Appendix), with all metrics except CKA achieving highly significant correlations ($p < 0.001$) for all three models. The correlation magnitudes are slightly lower than NER for AfriBERTa but comparable for AfroXLM-R and Serengeti. CKA again shows the same pattern: significant only for Serengeti.

\subsubsection{Sentiment Analysis}

Sentiment analysis shows markedly weaker correlations than NER and POS (Table~\ref{tab:sentiment-results-app}, Appendix). Only 2 of 15 metric-model combinations reach statistical significance (p < 0.05), likely due to smaller sample size (n = 30), domain mismatch between FLORES-200 and AfriSenti (Twitter), and task characteristics favoring lexical cues over embedding alignment.


With $n = 30$, we can reliably detect only $\rho > 0.36$ at $p = 0.05$; the observed correlations ($\rho = 0.12$--$0.38$) sit near this threshold, so weak effects are difficult to distinguish from noise.

\subsection{Metric Inter-Correlation Analysis}
\label{sec:inter-metric}

We computed Spearman correlations between metric values across all 396 language-pair-model combinations (Table~\ref{tab:inter-metric-app}, Appendix). Two findings stand out. First, P@1 and CSLS are nearly identical ($\rho = 0.97$); practitioners need only compute one. Second, cosine\_gap shows weak correlation with retrieval-based metrics ($\rho = 0.15$--$0.20$), indicating they capture complementary signals. CKA is essentially independent of cosine\_gap ($\rho = -0.02$).

\subsection{Domain Effect Analysis}
\label{sec:domain}



All metrics show weaker correlations for the Twitter-domain sentiment task compared to formal-text tasks (NER, POS), with cosine\_gap exhibiting the largest drop ($\Delta = -0.28$; see Table~\ref{tab:domain-app}, Appendix). This suggests that embedding metrics computed on formal text may transfer less reliably to social media domains, though the effect is confounded with task type and sample size.

\subsection{CKA Analysis}
\label{sec:cka}



CKA achieves significant correlation in only 2 of 9 conditions, both for Serengeti (NER and POS: $\rho = 0.38$; see Table~\ref{tab:cka-app}, Appendix).
Possible explanations are discussed in \S\ref{sec:discussion}.

\subsection{Model-Specific Patterns}
\label{sec:model-patterns}



AfroXLM-R shows consistently weaker correlations than AfriBERTa and Serengeti across all metrics and tasks (mean $\rho = 0.22$ vs. $0.37$--$0.39$; see Table~\ref{tab:model-summary-app}, Appendix). The source of this difference, whether architectural, related to pretraining data, or stemming from the adaptation process, is unclear and discussed in \S\ref{sec:discussion}. Practitioners using AfroXLM-R should expect weaker predictive signal from embedding similarity metrics.

\subsection{Effect of Pretraining Language Inclusion}
\label{sec:pretraining-effects}

Transfer to seen target languages yields 25--27\% higher F1 than transfer to unseen targets (Table~\ref{tab:pretraining-effect-app}, Appendix). Notably, target language inclusion matters more than source language inclusion. For metric validity, cosine\_gap remains a significant predictor for unseen targets with AfriBERTa ($\rho = 0.44$) but loses predictive power for AfroXLM-R ($\rho = 0.09$; Table~\ref{tab:metric-by-pretraining-app}, Appendix). This suggests practitioners can use embedding similarity to guide source selection even for target languages not in pretraining, though reliability varies by model.

\subsection{Comparison with Linguistic Features}
\label{sec:uriel}

Prior work has used linguistic typology features from URIEL \citep{littell-etal-2017-uriel} to predict cross-lingual transfer. We compare embedding metrics against URIEL genetic distance (Table~\ref{tab:uriel-combined}).



\begin{table}[t]
\centering
\small
\begin{tabular}{@{}llccc@{}}
\toprule
\textbf{Task} & \textbf{Feature} & \textbf{AfriBERTa} & \textbf{AfroXLM-R} & \textbf{Serengeti} \\
\midrule
\multirow{2}{*}{NER} & cosine\_gap & \textbf{0.60} & 0.37 & \textbf{0.51} \\
& URIEL genetic & 0.27 & \textbf{0.43} & 0.40 \\
\midrule
\multirow{2}{*}{POS} & cosine\_gap & \textbf{0.56} & 0.49 & \textbf{0.47} \\
& URIEL genetic & 0.45 & \textbf{0.65} & 0.45 \\
\bottomrule
\end{tabular}
\caption{Embedding metrics vs.\ URIEL genetic distance ($|\rho|$ shown). URIEL correlations are negative in sign (distance vs.\ performance); we report absolute values. Bold indicates higher $|\rho|$ per model.}
\label{tab:uriel-combined}
\end{table}

For NER, embedding metrics outperform URIEL for AfriBERTa ($|\rho| = 0.60$ vs. $0.27$) and Serengeti ($|\rho| = 0.51$ vs. $0.40$), but URIEL shows competitive performance for AfroXLM-R ($|\rho| = 0.43$ vs. $0.37$). For POS, the pattern is similar, with URIEL substantially outperforming embedding metrics for AfroXLM-R ($|\rho| = 0.65$ vs. $0.49$).

These results suggest that when embedding metrics show weak correlations (as with AfroXLM-R), linguistic distance features may provide a viable alternative. Combining embedding and linguistic features could potentially improve source selection, though we leave this exploration to future work.





\subsection{Practical Guidance}
\label{sec:practical}

For practitioners seeking to select source languages for cross-lingual transfer to African languages, we offer the following guidance based on our findings.

\textbf{Recommended approach:}
\begin{enumerate}
    \item Use cosine\_gap or P@1 as default metrics (they capture different but complementary signal).
    \item Always compute metrics separately for your specific model; do not pool across models.
    \item Do not rely on CKA alone.
    \item For AfroXLM-R, consider URIEL linguistic features as an alternative or supplement.
\end{enumerate}

\textbf{Expected performance:} To quantify practical utility, we evaluated source selection accuracy using cosine\_gap. For each target language, we ranked all candidate source languages by their cosine\_gap values and checked whether the empirically best source (highest transfer performance) appeared in the top-$K$ ranked positions. Table~\ref{tab:practical} reports accuracy across tasks and models.

\begin{table}[t]
\centering
\footnotesize
\begin{tabular}{@{}llcccc@{}}
\toprule
\textbf{Task} & \textbf{Model} & \textbf{Top-1} & \textbf{Top-3} & \textbf{Rand-1} & \textbf{Rand-3} \\
\midrule
\multirow{3}{*}{NER} & AfriBERTa & 50\% & 83\% & 9\% & 27\% \\
& AfroXLM-R & 25\% & 42\% & 9\% & 27\% \\
& Serengeti & 33\% & 42\% & 9\% & 27\% \\
\midrule
\multirow{3}{*}{POS} & AfriBERTa & 36\% & 55\% & 10\% & 30\% \\
& AfroXLM-R & 27\% & 45\% & 10\% & 30\% \\
& Serengeti & 36\% & 36\% & 10\% & 30\% \\
\midrule
\multirow{3}{*}{Sent.} & AfriBERTa & 33\% & 33\% & 20\% & 60\% \\
& AfroXLM-R & 33\% & 50\% & 20\% & 60\% \\
& Serengeti & 17\% & 17\% & 20\% & 60\% \\
\bottomrule
\end{tabular}
\caption{Source selection accuracy using cosine\_gap. \textbf{Top-$K$}: percentage of target languages for which the empirically best source appears in the top-$K$ ranked by cosine\_gap. \textbf{Rand-$K$}: random baseline ($K/(N-1)$ where $N$ is the number of languages and $N-1$ is the number of candidate sources per target). For each target language, we ranked the $N-1$ candidate source languages by cosine\_gap and checked if the oracle best appeared in the top-$K$.}
\label{tab:practical}
\end{table}

For NER with AfriBERTa, cosine\_gap correctly identifies the best source language 50\% of the time (vs. 9\% random), and places it in the top 3 candidates 83\% of the time (vs. 27\% random). Performance is weaker but still above random for other model-task combinations, except Sentiment with Serengeti, where metric-based selection performs below random chance.


\textbf{Caveats:} Expect weaker signal for AfroXLM-R and for informal domains (e.g., Twitter). Metrics provide guidance, not guarantees; validate top candidates experimentally and consider URIEL if embedding metrics show weak signal.

\section{Discussion}
\label{sec:discussion}

\subsection{Interpreting the Results}
\label{sec:interpreting}


{\bf CKA.} CKA achieves significance in only 2 of 9 conditions (both Serengeti), indicating that CKA-style structural similarity is not a robust proxy for transfer in our setting; practitioners should prefer sentence-level metrics.

{\bf AfroXLM-R.} AfroXLM-R shows weaker metric-transfer correlations (mean $\rho = 0.22$ vs.\ $0.37$--$0.39$). URIEL features outperform embedding metrics for this model in our experiments, suggesting typology-based selection may be preferable for adapted models.

{\bf Sentiment.} Sentiment shows weaker correlations than NER and POS (\S\ref{sec:task-results}). Given domain mismatch (FLORES-200 vs.\ Twitter) and small sample size ($n = 30$), we treat these results as inconclusive.

{\bf Simpson's Paradox.} Correlation signs reverse when pooling across models ($\rho = -0.18$ pooled vs.\ $0.37$--$0.60$ stratified), highlighting that analyses must stratify by model and pretraining regime. These findings motivate the workflow in \S\ref{sec:practical}.

\subsection{Future Work}
\label{sec:future-work}



Key directions include (i) testing directional CKA variants to explain its failure, (ii) analyzing how continued pretraining reshapes AfroXLM-R's language-space geometry, (iii) replicating on larger sentiment datasets and additional architectures (decoder-only, encoder-decoder), and (iv) combining embedding metrics with URIEL and evaluating multi-source transfer.

\subsection{Broader Impact}
\label{sec:broader-impact}


Better source selection can reduce annotation burden for African languages with limited resources. However, embedding metrics achieve only moderate predictive power; practitioners should use them as guidance alongside expert judgment.

\section{Conclusion}
\label{sec:conclusion}

We investigated whether embedding similarity metrics can predict cross-lingual transfer for African languages. Our experiments across three models, three tasks, and 12 languages reveal that cosine\_gap, P@1, and CSLS achieve moderate correlations ($\rho \approx 0.40$), substantially outperforming random selection (50\% vs.\ 9\% top-1 accuracy for NER with AfriBERTa), while CKA proves unreliable.

Critically, model stratification is essential: pooling across models produces Simpson's Paradox, yielding misleading negative correlations ($\rho = -0.18$) that mask positive within-model relationships ($\rho = 0.37$--$0.60$). We recommend cosine\_gap as a first-pass filter, validating top candidates experimentally; when embedding metrics show weak signal (particularly for AfroXLM-R), URIEL features provide an alternative.

Our findings are specific to encoder-only models and discriminative tasks. To our knowledge, this work provides the first systematic comparison of embedding metrics for African language transfer, offering practical guidance and methodological insights for multilingual NLP.

\section*{Limitations}
\label{sec:limitations}

We acknowledge several limitations of this study.

{\bf Limited Model Coverage.}
We tested only three encoder-only multilingual models (AfriBERTa, AfroXLM-R, Serengeti). Results may not generalize to decoder-only models (GPT-style architectures, BLOOM), encoder-decoder models (mT5, mBART), or larger-scale models. The cross-lingual alignment properties of decoder-only architectures differ fundamentally from encoder models, and our findings should not be assumed to transfer.

{\bf Limited Task Coverage.}
We evaluated NER, POS tagging, and sentiment classification, all token or sequence classification tasks. Other tasks such as question answering, machine translation, and summarization may show different patterns. Generation tasks in particular may depend on different aspects of cross-lingual alignment.

{\bf Metric Computation Scope.}
All embedding metrics were computed from FLORES-200, a curated parallel evaluation corpus in a formal written style with approximately 1,000 sentences per language. This limits our findings to languages covered by FLORES-200. Domain-specific metrics computed on Twitter or other parallel data might reveal different patterns.

{\bf Correlation, Not Causation.}
We demonstrate that embedding metrics correlate with transfer performance, not that they cause good transfer. Underlying factors such as linguistic similarity, pretraining data overlap, or shared vocabulary may drive both metric values and transfer performance. Metrics are predictive proxies, not causal mechanisms.

{\bf Language Family Scope.}
Our experiments covered 12 African languages spanning Niger-Congo, Afro-Asiatic, and Nilo-Saharan families. Patterns may differ for other low-resource language families (Austronesian, Dravidian, Indigenous American languages) with different typological properties or for high-resource to low-resource transfer scenarios (e.g., English to African languages).

\section*{Ethics Statement}
Our study uses publicly available pretrained models and datasets. We do not introduce new data collection or human annotation. While improved transfer prediction can help expand NLP support for under-resourced languages, model errors may still disproportionately affect speakers of low-resource languages. We report limitations and recommend validation before deployment.



\bibliography{custom}

@article{adelani-etal-2021-masakhaner,
    title = "{M}asakha{NER}: Named Entity Recognition for {A}frican Languages",
    author = "Adelani, David and Abbott, Jade and Neubig, Graham and D{'}souza, Daniel and Kreutzer, Julia and Lignos, Constantine and Palen-Michel, Chester and Buzaaba, Happy and Rijhwani, Shruti and Ruder, Sebastian and others",
    journal = "Transactions of the Association for Computational Linguistics",
    volume = "9",
    year = "2021",
    pages = "1116--1131",
}

@inproceedings{adelani-etal-2022-masakhaner,
    title = "{M}asakha{NER} 2.0: {A}frica-centric Transfer Learning for Named Entity Recognition",
    author = "Adelani, David and Neubig, Graham and Ruder, Sebastian and Rijhwani, Shruti and Beukman, Michael and Palen-Michel, Chester and Lignos, Constantine and Alabi, Jesujoba and Muhammad, Shamsuddeen and Nabende, Peter and others",
    booktitle = "Proceedings of the 2022 Conference on Empirical Methods in Natural Language Processing",
    year = "2022",
    pages = "4488--4508",
}

@inproceedings{dione-etal-2023-masakhapos,
    title = "{M}asakha{POS}: Part-of-Speech Tagging for Typologically Diverse {A}frican Languages",
    author = "Dione, Cheikh M. Bamba and Adelani, David Ifeoluwa and Nabende, Peter and Alabi, Jesujoba Oluwadara and Sindane, Thapelo and Buzaaba, Happy and Muhammad, Shamsuddeen Hassan and Emezue, Chris C. and Ogayo, Perez and Aremu, Anuoluwapo and others",
    booktitle = "Proceedings of the 61st Annual Meeting of the Association for Computational Linguistics (Volume 1: Long Papers)",
    year = "2023",
    pages = "1763--1801",
    publisher = "Association for Computational Linguistics",
}

@inproceedings{muhammad-etal-2023-afrisenti,
    title = "{A}fri{S}enti: A {T}witter Sentiment Analysis Benchmark for {A}frican Languages",
    author = "Muhammad, Shamsuddeen Hassan and Adelani, David and others",
    booktitle = "Proceedings of the 2023 Conference on Empirical Methods in Natural Language Processing",
    year = "2023",
    pages = "13968--13981",
    publisher = "Association for Computational Linguistics",
}

@inproceedings{devlin-etal-2019-bert,
    title = "{BERT}: Pre-training of Deep Bidirectional Transformers for Language Understanding",
    author = "Devlin, Jacob and Chang, Ming-Wei and Lee, Kenton and Toutanova, Kristina",
    booktitle = "Proceedings of the 2019 Conference of the North {A}merican Chapter of the Association for Computational Linguistics: Human Language Technologies",
    year = "2019",
    pages = "4171--4186",
}

@inproceedings{conneau-etal-2020-unsupervised,
    title = "Unsupervised Cross-lingual Representation Learning at Scale",
    author = "Conneau, Alexis and Khandelwal, Kartikay and Goyal, Naman and Chaudhary, Vishrav and Wenzek, Guillaume and Guzm{\'a}n, Francisco and Grave, Edouard and Ott, Myle and Zettlemoyer, Luke and Stoyanov, Veselin",
    booktitle = "Proceedings of the 58th Annual Meeting of the Association for Computational Linguistics",
    year = "2020",
    pages = "8440--8451",
}

@inproceedings{ogueji-etal-2021-small,
    title = "Small Data? No Problem! Exploring the Viability of Pretrained Multilingual Language Models for Low-resourced Languages",
    author = "Ogueji, Kelechi and Zhu, Yuxin and Lin, Jimmy",
    booktitle = "Proceedings of the 1st Workshop on Multilingual Representation Learning",
    year = "2021",
    pages = "116--126",
}

@inproceedings{alabi-etal-2022-adapting,
    title = "Adapting Pre-trained Language Models to {A}frican Languages via Multilingual Adaptive Fine-Tuning",
    author = "Alabi, Jesujoba and Adelani, David and Mosbach, Marius and Klakow, Dietrich",
    booktitle = "Proceedings of the 29th International Conference on Computational Linguistics",
    year = "2022",
    pages = "4336--4349",
}

@inproceedings{adebara-etal-2023-serengeti,
    title = "{SERENGETI}: Massively Multilingual Language Models for {A}frica",
    author = "Adebara, Ife and Elmadany, AbdelRahim and Abdul-Mageed, Muhammad and Alcoba Inciarte, Guillem",
    booktitle = "Findings of the Association for Computational Linguistics: ACL 2023",
    year = "2023",
    pages = "1498--1537",
}

@inproceedings{lin-etal-2019-choosing,
    title = "Choosing Transfer Languages for Cross-Lingual Learning",
    author = "Lin, Yu-Hsiang and Chen, Chian-Yu and Lee, Jean and Li, Zirui and Zhang, Yuyan and Xia, Mengzhou and Rijhwani, Shruti and He, Junxian and Zhang, Zhisong and Ma, Xuezhe and Anastasopoulos, Antonios and Littell, Patrick and Neubig, Graham",
    booktitle = "Proceedings of the 57th Annual Meeting of the Association for Computational Linguistics",
    year = "2019",
    pages = "3125--3135",
}

@inproceedings{pires-etal-2019-multilingual,
    title = "How Multilingual is Multilingual {BERT}?",
    author = "Pires, Telmo and Schlinger, Eva and Garrette, Dan",
    booktitle = "Proceedings of the 57th Annual Meeting of the Association for Computational Linguistics",
    year = "2019",
    pages = "4996--5007",
    publisher = "Association for Computational Linguistics",
}

@inproceedings{littell-etal-2017-uriel,
    title = "{URIEL} and lang2vec: Representing languages as typological, geographical, and phylogenetic vectors",
    author = "Littell, Patrick and Mortensen, David R. and Lin, Ke and Kairis, Katherine and Turner, Carlisle and Levin, Lori",
    booktitle = "Proceedings of the 15th Conference of the {E}uropean Chapter of the Association for Computational Linguistics: Volume 2, Short Papers",
    year = "2017",
    pages = "8--14",
}

@inproceedings{muller-etal-2023-languages,
    title = "Languages You Know Influence Those You Learn: Impact of Language Characteristics on Multi-Lingual Text-to-Text Transfer",
    author = "Muller, Benjamin and Gupta, Deepanshu and Patwardhan, Siddharth and Fauconnier, Jean-Philippe and Vandyke, David and Agarwal, Sachin",
    booktitle = "Proceedings of Machine Learning Research",
    volume = "203",
    year = "2023",
    publisher = "PMLR",
}

@inproceedings{kornblith2019similarity,
    title = "Similarity of Neural Network Representations Revisited",
    author = "Kornblith, Simon and Norouzi, Mohammad and Lee, Honglak and Hinton, Geoffrey",
    booktitle = "Proceedings of the 36th International Conference on Machine Learning",
    year = "2019",
    pages = "3519--3529",
}

@inproceedings{ethayarajh-2019-contextual,
    title = "How Contextual are Contextualized Word Representations? {C}omparing the Geometry of {BERT}, {ELMo}, and {GPT}-2 Embeddings",
    author = "Ethayarajh, Kawin",
    booktitle = "Proceedings of the 2019 Conference on Empirical Methods in Natural Language Processing and the 9th International Joint Conference on Natural Language Processing ({EMNLP}-{IJCNLP})",
    year = "2019",
    pages = "55--65",
    publisher = "Association for Computational Linguistics",
}

@inproceedings{gao-etal-2021-simcse,
    title = "{S}im{CSE}: Simple Contrastive Learning of Sentence Embeddings",
    author = "Gao, Tianyu and Yao, Xingcheng and Chen, Danqi",
    booktitle = "Proceedings of the 2021 Conference on Empirical Methods in Natural Language Processing",
    year = "2021",
    pages = "6894--6910",
    publisher = "Association for Computational Linguistics",
}

@inproceedings{conneau-etal-2017-word,
    title = "Word Translation Without Parallel Data",
    author = "Conneau, Alexis and Lample, Guillaume and Ranzato, Marc{'}Aurelio and Denoyer, Ludovic and J{\'e}gou, Herv{\'e}",
    booktitle = "Proceedings of the 5th International Conference on Learning Representations",
    year = "2017",
}

@inproceedings{del-fishel-2022-cross,
    title = "Cross-lingual Similarity of Multilingual Representations Revisited",
    author = "Del, Maksym and Fishel, Mark",
    booktitle = "Proceedings of the 2nd Conference of the Asia-Pacific Chapter of the Association for Computational Linguistics and the 12th International Joint Conference on Natural Language Processing (Volume 1: Long Papers)",
    year = "2022",
    pages = "185--195",
    publisher = "Association for Computational Linguistics",
}

@inproceedings{rajaee-pilehvar-2021-cluster,
    title = "A Cluster-based Approach for Improving Isotropy in Contextual Embedding Space",
    author = "Rajaee, Sara and Pilehvar, Mohammad Taher",
    booktitle = "Proceedings of the 59th Annual Meeting of the Association for Computational Linguistics and the 11th International Joint Conference on Natural Language Processing (Volume 2: Short Papers)",
    year = "2021",
    pages = "575--582",
    publisher = "Association for Computational Linguistics",
}

@inproceedings{kargaran-etal-2025-mexa,
    title = "{MEXA}: Multilingual Evaluation of {E}nglish-Centric {LLM}s via Cross-Lingual Alignment",
    author = "Kargaran, Amir Hossein and Modarressi, Ali and Nikeghbal, Nafiseh and Diesner, Jana and Yvon, Fran{\c{c}}ois and Sch{\"u}tze, Hinrich",
    booktitle = "Findings of the Association for Computational Linguistics: ACL 2025",
    year = "2025",
    pages = "27001--27023",
    publisher = "Association for Computational Linguistics",
}

@inproceedings{wang-etal-2024-probing-emergence,
    title = "Probing the Emergence of Cross-lingual Alignment during {LLM} Training",
    author = "Wang, Hetong and Minervini, Pasquale and Ponti, Edoardo",
    booktitle = "Findings of the Association for Computational Linguistics: ACL 2024",
    year = "2024",
    pages = "12159--12173",
    publisher = "Association for Computational Linguistics",
}

@inproceedings{zhao-etal-2023-joint,
    title = "A Joint Matrix Factorization Analysis of Multilingual Representations",
    author = "Zhao, Zheng and Ziser, Yftah and Webber, Bonnie and Cohen, Shay",
    booktitle = "Findings of the Association for Computational Linguistics: EMNLP 2023",
    year = "2023",
    pages = "12764--12783",
    publisher = "Association for Computational Linguistics",
}

@inproceedings{ravisankar-etal-2025-map,
    title = "Can you map it to {E}nglish? The Role of Cross-Lingual Alignment in Multilingual Performance of {LLM}s",
    author = "Ravisankar, Kartik and Han, Hyojung and Carpuat, Marine",
    booktitle = "arXiv preprint arXiv:2504.09378",
    year = "2025",
}

@inproceedings{litschko-etal-2020-evaluating,
    title = "Evaluating Multilingual Text Encoders for Unsupervised Cross-Lingual Retrieval",
    author = "Litschko, Robert and Glava{\v{s}}, Goran and Ponzetto, Simone Paolo and Vuli{\'c}, Ivan",
    booktitle = "European Conference on Information Retrieval",
    year = "2020",
    publisher = "Springer",
}

@inproceedings{hedderich-etal-2020-transfer,
    title = "Transfer Learning and Distant Supervision for Multilingual Transformer Models: {A} Study on {A}frican Languages",
    author = "Hedderich, Michael A and Adelani, David Ifeoluwa and Zhu, Dawei and Alabi, Jesujoba Oluwadara and Markus, Udia and Klakow, Dietrich",
    booktitle = "Proceedings of the 2020 Conference on Empirical Methods in Natural Language Processing ({EMNLP})",
    year = "2020",
    pages = "2691--2707",
    publisher = "Association for Computational Linguistics",
}

@inproceedings{adelani-etal-2022-thousand,
    title = "A Few Thousand Translations Go a Long Way! {L}everaging Pre-trained Models for {A}frican News Translation",
    author = "Adelani, David Ifeoluwa and Alabi, Jesujoba Oluwadara and Fan, Angela and Kreutzer, Julia and Shen, Xiaoyu and Reid, Machel and Ruiter, Dana and Klakow, Dietrich and Nabende, Peter and Chang, Ernie and others",
    booktitle = "Proceedings of the 2022 Conference of the North American Chapter of the Association for Computational Linguistics: Human Language Technologies",
    year = "2022",
    pages = "37--57",
    publisher = "Association for Computational Linguistics",
}

@inproceedings{adebara-abdul-mageed-2022-towards,
    title = "Towards {A}frocentric {NLP} for {A}frican Languages: {W}here We {A}re and Where We {S}hould {G}o",
    author = "Adebara, Ife and Abdul-Mageed, Muhammad",
    booktitle = "Findings of the Association for Computational Linguistics: ACL 2023",
    year = "2023",
    pages = "1533--1547",
    publisher = "Association for Computational Linguistics",
}

@inproceedings{rice-etal-2025-typology,
    title = "Untangling the Influence of Typology, Data and Model Architecture on Ranking Transfer Languages for Cross-Lingual {POS} Tagging",
    author = "Rice, Enora and Marashian, Ali and Haynie, Hannah and Wense, K. and Palmer, Alexis",
    booktitle = "Proceedings of the 16th Conference of the European Chapter of the Association for Computational Linguistics",
    year = "2025",
    publisher = "Association for Computational Linguistics",
}

@inproceedings{conneau-etal-2018-xnli,
    title = "{XNLI}: Evaluating Cross-lingual Sentence Representations",
    author = "Conneau, Alexis and Rinott, Ruty and Lample, Guillaume and Williams, Adina and Bowman, Samuel and Schwenk, Holger and Stoyanov, Veselin",
    booktitle = "Proceedings of the 2018 Conference on Empirical Methods in Natural Language Processing",
    year = "2018",
    pages = "2475--2485",
    publisher = "Association for Computational Linguistics",
}

@inproceedings{artetxe-schwenk-2019-massively,
    title = "Massively Multilingual Sentence Embeddings for Zero-Shot Cross-Lingual Transfer and Beyond",
    author = "Artetxe, Mikel and Schwenk, Holger",
    booktitle = "Proceedings of the 57th Annual Meeting of the Association for Computational Linguistics",
    year = "2019",
    pages = "52--68",
}

@article{nllb2024,
  author = {{NLLB Team}},
  title = {Scaling neural machine translation to 200 languages},
  journal = {Nature},
  volume = {630},
  pages = {841--846},
  year = {2024},
  doi = {10.1038/s41586-024-07335-x}
}

@article{song2012feature,
    title = "Feature Selection via Dependence Maximization",
    author = "Song, Le and Smola, Alex and Gretton, Arthur and Bedo, Justin and Borgwardt, Karsten M.",
    journal = "Journal of Machine Learning Research",
    volume = "13",
    pages = "1393--1434",
    year = "2012",
    doi = "10.1162/jmlr.2012.13.may.1393"
}

@inproceedings{mosbach-etal-2021-stability,
    title = "On the Stability of Fine-tuning {BERT}: Misconceptions, Explanations, and Strong Baselines",
    author = "Mosbach, Marius and Andriushchenko, Maksym and Klakow, Dietrich",
    booktitle = "International Conference on Learning Representations",
    year = "2021",
    url = "https://openreview.net/forum?id=nzpLWnVAyah",
}

@inproceedings{lauscher-etal-2020-zero,
    title = "From Zero to Hero: On the Limitations of Zero-Shot Cross-Lingual Transfer with Multilingual Transformers",
    author = {Lauscher, Anne and Ravishankar, Vinit and Vuli\'{c}, Ivan and Glava\v{s}, Goran},
    booktitle = "Proceedings of the 2020 Conference on Empirical Methods in Natural Language Processing (EMNLP)",
    year = "2020",
    pages = "4547--4563",
    publisher = "Association for Computational Linguistics",
    doi = "10.18653/v1/2020.emnlp-main.363"
}

@inproceedings{turc-etal-2021-revisiting,
    title = "Revisiting the Primacy of {E}nglish in Zero-shot Cross-lingual Transfer",
    author = "Turc, Iulia and Lee, Kenton and Eisenstein, Jacob and Chang, Ming-Wei and Toutanova, Kristina",
    booktitle = "Proceedings of the 2nd Workshop on Evaluation and Comparison of NLP Systems",
    year = "2021",
    pages = "149--162",
    publisher = "Association for Computational Linguistics",
    doi = "10.18653/v1/2021.eval4nlp-1.17"
}

@article{simpson1951interpretation,
    title = "The Interpretation of Interaction in Contingency Tables",
    author = "Simpson, Edward H.",
    journal = "Journal of the Royal Statistical Society: Series B (Methodological)",
    volume = "13",
    number = "2",
    pages = "238--241",
    year = "1951",
}

@incollection{creissels-2019-morphology-niger-congo,
    author = {Creissels, Denis},
    title = {Morphology in {Niger-Congo} Languages},
    booktitle = {Oxford Research Encyclopedia of Linguistics},
    publisher = {Oxford University Press},
    year = {2019},
    month = {05},
    doi = {10.1093/acrefore/9780199384655.013.535},
    url = {https://oxfordre.com/linguistics/view/10.1093/acrefore/9780199384655.001.0001/acrefore-9780199384655-e-535}
}

@incollection{hyman-etal-2019-niger-congo,
    author = {Hyman, Larry M. and Rolle, Nicholas and Sande, Hannah and Clem, Emily and Jenks, Peter S. E. and Lionnet, Florian and Merrill, John and Baier, Nicholas},
    title = {{Niger-Congo} Linguistic Features and Typology},
    booktitle = {The Cambridge Handbook of African Linguistics},
    editor = {Wolff, H. Ekkehard},
    publisher = {Cambridge University Press},
    address = {Cambridge},
    year = {2019},
    pages = {191--245},
    series = {Cambridge Handbooks in Language and Linguistics}
}

@inproceedings{muller-etal-2021-first,
    title = "First Align, then Predict: Understanding the Cross-Lingual Ability of Multilingual {BERT}",
    author = "M{\"u}ller, Benjamin and
      Elazar, Yanai and
      Sagot, Beno{\^i}t and
      Seddah, Djam{\'e}",
    booktitle = "Proceedings of the 16th Conference of the European Chapter of the Association for Computational Linguistics: Main Volume",
    year = "2021",
    address = "Online",
    publisher = "Association for Computational Linguistics",
    pages = "2214--2231",
}

@inproceedings{conneau-etal-2020-emerging,
    title = "Emerging Cross-lingual Structure in Pretrained Language Models",
    author = "Conneau, Alexis and Wu, Shijie and Li, Haoran and Zettlemoyer, Luke and Stoyanov, Veselin",
    booktitle = "Proceedings of the 58th Annual Meeting of the Association for Computational Linguistics",
    year = "2020",
    address = "Online",
    publisher = "Association for Computational Linguistics",
    pages = "6022--6034",
}

\clearpage
\appendix

\section{Pretraining Language Coverage}
\label{app:pretraining}

Table~\ref{tab:lang-coverage-app} shows which experimental languages appear in each model's pretraining data.

\begin{table}[H]
\centering
\small
\begin{tabular}{lccc}
\toprule
\textbf{Language} & \textbf{AfriBERTa} & \textbf{AfroXLM-R} & \textbf{Serengeti} \\
\midrule
Amharic & \checkmark & \checkmark & \checkmark \\
Bambara & -- & -- & \checkmark \\
Hausa & \checkmark & \checkmark & \checkmark \\
Igbo & \checkmark & \checkmark & \checkmark \\
Kinyarwanda & \checkmark$^\dagger$ & \checkmark & \checkmark \\
Luganda & -- & \checkmark & \checkmark \\
Luo & -- & \checkmark & \checkmark \\
Swahili & \checkmark & \checkmark & \checkmark \\
Wolof & -- & \checkmark & \checkmark \\
Xhosa & -- & -- & \checkmark \\
Yoruba & \checkmark & \checkmark & \checkmark \\
Zulu & -- & -- & \checkmark \\
\midrule
\textbf{Coverage} & 6/12 & 9/12 & 12/12 \\
\bottomrule
\end{tabular}
\caption{Pretraining language coverage for experimental languages. \checkmark\ = included in pretraining; -- = not included. $^\dagger$AfriBERTa includes Gahuza, a mixed Kinyarwanda/Kirundi corpus.}
\label{tab:lang-coverage-app}
\end{table}

\section{Training Configuration}
\label{app:hyperparams}

Table~\ref{tab:hyperparams-app} details the hyperparameters used for fine-tuning.

\begin{table}[H]
\centering
\small
\begin{tabular}{@{}ll@{}}
\toprule
\textbf{Parameter} & \textbf{Value} \\
\midrule
Learning rate & $2 \times 10^{-5}$ \\
LR schedule & Linear warmup (10\%) + decay \\
Optimizer & AdamW with bias correction \\
Batch size & 16 \\
Max epochs & 10 \\
Early stopping & Patience = 3 \\
Max sequence length & 128 tokens \\
Dropout & 0.1 \\
\bottomrule
\end{tabular}
\caption{Fine-tuning hyperparameters. Learning rate and schedule follow \citet{devlin-etal-2019-bert}; bias correction follows \citet{mosbach-etal-2021-stability}.}
\label{tab:hyperparams-app}
\end{table}

\section{Task-Specific Correlation Results}
\label{app:task-correlations}

Tables~\ref{tab:ner-results-app}--\ref{tab:sentiment-results-app} present detailed correlations between embedding metrics and transfer performance for each task.

\begin{table}[H]
\centering
\small
\begin{tabular}{@{}lccc@{}}
\toprule
\textbf{Metric} & \textbf{AfriBERTa} & \textbf{AfroXLM-R} & \textbf{Serengeti} \\
\midrule
cosine\_mean & 0.35*** & 0.10 & 0.52*** \\
cosine\_gap & 0.60*** & 0.37*** & 0.51*** \\
P@1 & 0.56*** & 0.20* & 0.53*** \\
CSLS & 0.58*** & 0.24** & 0.52*** \\
CKA & 0.03 & $-$0.04 & 0.38*** \\
\bottomrule
\end{tabular}
\caption{Spearman correlations between embedding metrics and NER transfer F1 ($n = 132$ pairs per model). Significance: *** $p < 0.001$, ** $p < 0.01$, * $p < 0.05$.}
\label{tab:ner-results-app}
\end{table}

\begin{table}[H]
\centering
\small
\begin{tabular}{@{}lccc@{}}
\toprule
\textbf{Metric} & \textbf{AfriBERTa} & \textbf{AfroXLM-R} & \textbf{Serengeti} \\
\midrule
cosine\_mean & 0.49*** & 0.27** & 0.53*** \\
cosine\_gap & 0.56*** & 0.49*** & 0.47*** \\
P@1 & 0.50*** & 0.43*** & 0.50*** \\
CSLS & 0.50*** & 0.41*** & 0.53*** \\
CKA & 0.16 & $-$0.13 & 0.38*** \\
\bottomrule
\end{tabular}
\caption{Spearman correlations between embedding metrics and POS transfer accuracy ($n = 110$ pairs per model). Significance: *** $p < 0.001$, ** $p < 0.01$, * $p < 0.05$.}
\label{tab:pos-results-app}
\end{table}

\begin{table}[H]
\centering
\small
\begin{tabular}{@{}lccc@{}}
\toprule
\textbf{Metric} & \textbf{AfriBERTa} & \textbf{AfroXLM-R} & \textbf{Serengeti} \\
\midrule
cosine\_mean & 0.38* & 0.12 & 0.26 \\
cosine\_gap & 0.32 & 0.16 & 0.19 \\
P@1 & 0.37* & 0.24 & 0.26 \\
CSLS & 0.30 & 0.23 & 0.25 \\
CKA & 0.00 & 0.11 & 0.05 \\
\bottomrule
\end{tabular}
\caption{Spearman correlations between embedding metrics and sentiment transfer F1 ($n = 30$ pairs per model). Significance: *** $p < 0.001$, ** $p < 0.01$, * $p < 0.05$.}
\label{tab:sentiment-results-app}
\end{table}

\section{Metric Inter-Correlations}
\label{app:inter-metric}

Table~\ref{tab:inter-metric-app} shows correlations between metric values (not with transfer performance).

\begin{table}[H]
\centering
\small
\begin{tabular}{lcc}
\toprule
\textbf{Metric Pair} & \textbf{$\rho$} & \textbf{Interpretation} \\
\midrule
P@1 vs CSLS & 0.97 & Nearly identical \\
cosine\_gap vs P@1 & 0.20 & Different signals \\
cosine\_gap vs CSLS & 0.15 & Different signals \\
cosine\_gap vs CKA & $-$0.02 & Independent \\
cosine\_mean vs cosine\_gap & $-$0.72 & Inversely related \\
\bottomrule
\end{tabular}
\caption{Inter-metric Spearman correlations computed across all 396 language-pair-model combinations.}
\label{tab:inter-metric-app}
\end{table}

\section{Domain Effects}
\label{app:domain}

Table~\ref{tab:domain-app} compares correlations for formal-text tasks versus Twitter-domain task.

\begin{table}[H]
\centering
\small
\begin{tabular}{@{}lccc@{}}
\toprule
\textbf{Metric} & \textbf{Formal} & \textbf{Twitter} & \textbf{$\Delta$} \\
\midrule
cosine\_mean & 0.38 & 0.26 & $-$0.12 \\
cosine\_gap & 0.50 & 0.22 & $-$0.28 \\
P@1 & 0.46 & 0.29 & $-$0.17 \\
CSLS & 0.46 & 0.26 & $-$0.21 \\
CKA & 0.13 & 0.05 & $-$0.07 \\
\bottomrule
\end{tabular}
\caption{Mean correlations for formal-text tasks (NER, POS; 6 conditions) versus Twitter-domain task (Sentiment; 3 conditions).}
\label{tab:domain-app}
\end{table}

\section{CKA Analysis}
\label{app:cka}

Table~\ref{tab:cka-app} isolates CKA performance across tasks.

\begin{table}[H]
\centering
\small
\begin{tabular}{lccc}
\toprule
\textbf{Task} & \textbf{AfriBERTa} & \textbf{AfroXLM-R} & \textbf{Serengeti} \\
\midrule
NER & 0.03 & $-$0.04 & 0.38*** \\
POS & 0.16 & $-$0.13 & 0.38*** \\
Sentiment & 0.00 & 0.11 & 0.05 \\
\bottomrule
\end{tabular}
\caption{CKA correlations with transfer performance. Significant results appear only for Serengeti (NER and POS).}
\label{tab:cka-app}
\end{table}

\section{Model-Level Summary}
\label{app:model-summary}

Table~\ref{tab:model-summary-app} summarizes correlation performance by model.

\begin{table}[H]
\centering
\small
\begin{tabular}{lccc}
\toprule
\textbf{Model} & \textbf{Mean $\rho$} & \textbf{Best} & \textbf{Worst} \\
\midrule
Serengeti & 0.39 & 0.53 & 0.05 \\
AfriBERTa & 0.37 & 0.60 & 0.00 \\
AfroXLM-R & 0.22 & 0.49 & $-$0.13 \\
\bottomrule
\end{tabular}
\caption{Model-level summary of metric-transfer correlations (averaged across 15 metric-task combinations per model).}
\label{tab:model-summary-app}
\end{table}

\section{Pretraining Effects}
\label{app:pretraining-effects}

Table~\ref{tab:pretraining-effect-app} compares transfer performance for seen versus unseen target languages.

\begin{table}[H]
\centering
\small
\begin{tabular}{@{}lccc@{}}
\toprule
\textbf{Model} & \textbf{Seen} & \textbf{Unseen} & \textbf{$\Delta$} \\
\midrule
AfriBERTa & 0.40 (n=66) & 0.31 (n=66) & +0.09 (+27\%) \\
AfroXLM-R & 0.58 (n=99) & 0.47 (n=33) & +0.12 (+25\%) \\
Serengeti & 0.52 (n=132) & -- & -- \\
\bottomrule
\end{tabular}
\caption{Transfer F1 by target language pretraining status. ``Seen'' = target language in model's pretraining; ``Unseen'' = not included. Serengeti has no unseen targets.}
\label{tab:pretraining-effect-app}
\end{table}

Table~\ref{tab:metric-by-pretraining-app} shows how metric validity varies by target language pretraining status.

\begin{table}[H]
\centering
\small
\begin{tabular}{lcc}
\toprule
\textbf{Model} & \textbf{Target Seen ($\rho$)} & \textbf{Target Unseen ($\rho$)} \\
\midrule
AfriBERTa & 0.68*** & 0.44*** \\
AfroXLM-R & 0.44*** & 0.09 \\
Serengeti & 0.51*** & -- \\
\bottomrule
\end{tabular}
\caption{Cosine\_gap correlation with transfer F1, stratified by target language pretraining status.}
\label{tab:metric-by-pretraining-app}
\end{table}

\end{document}